\documentclass[letterpaper, 10 pt, conference]{ieeeconf}
\usepackage{ifpdf}
\usepackage{amsmath,amsfonts}
\usepackage{amssymb}  
\usepackage{algorithmic}
\usepackage{algorithm}
\usepackage{array}
\usepackage{multirow}

\usepackage[caption=false,font=footnotesize]{subfig}
\usepackage{textcomp}
\usepackage{stfloats}
\usepackage{url}
\usepackage{verbatim}
\usepackage{soul}
\usepackage{graphicx}
\usepackage{cite}
\usepackage{booktabs}
\usepackage[export]{adjustbox}
\usepackage{bm}
\usepackage{nccmath}
\usepackage{mathtools}
\usepackage{tensor}
\usepackage[dvipsnames]{xcolor}
\usepackage{units}
\usepackage{tikz-cd}
\usepackage{bbm}
\usepackage{tikz}
\usepackage{cases}

\newcommand{\R}{\mathbb{R}}

\usepackage{hyperref}

\begin{document}

\title{Dynamic Loco-Manipulation via Hybrid Whole-body Bilateral Teleoperation of a Wheeled Humanoid}
\title{Whole-body Bilateral Teleoperation of a Wheeled Humanoid via Hybrid Switched Mappings and Moment Feedback for Dynamic Loco-Manipulation}
\title{Dynamic Loco-Manipulation of a Wheeled Humanoid via Bilateral Teleoperation with Position-Force Control Modes}
\title{Wheeled Humanoid Bilateral Teleoperation with Position-Force Control Modes for Dynamic Loco-Manipulation}

\author{{Amartya Purushottam$^1$, Jack Yan$^{3}$, Christopher Xu$^1$, Youngwoo Sim$^3$ and Joao Ramos$^{1,3}$}
\thanks{Manuscript received: July 3, 2023; Revised September 27, 2023; Accepted October 28, 2023.}
\thanks{This paper was recommended for publication by
Editor Jee-Hwan Ryu upon evaluation of the Associate Editor and Reviewers’
comments.}
\thanks{$^{1,2}$The authors are with the $^1$Department of Electrical and Computer Engineering and the $^2$Department of Mechanical Science and Engineering at the University of Illinois at Urbana-Champaign, USA.}
\thanks{This work is supported by the National Science Foundation via grant IIS-2024775.}
\thanks{Digital Object Identifier (DOI): see top of this page.}}

\maketitle

\begin{abstract}

Remote-controlled humanoid robots can revolutionize manufacturing, construction, and healthcare industries by performing complex or dangerous manual tasks traditionally done by humans. We refer to these behaviors as Dynamic Loco-Manipulation (DLM). To successfully complete these tasks, humans control the position of their bodies and contact forces at their hands. To enable similar whole-body control in humanoids, we introduce loco-manipulation retargeting strategies with switched position and force control modes in a bilateral teleoperation framework. Our proposed locomotion mappings use the pitch and yaw of the operator's torso to control robot position or acceleration. The manipulation retargeting maps the operator's arm movements to the robot's arms for joint-position or impedance control of the end-effector. A Human-Machine Interface captures the teleoperator's motion and provides haptic feedback to their torso, enhancing their awareness of the robot's interactions with the environment. In this paper, we demonstrate two forms of DLM. First, we show the robot slotting heavy boxes ($5-10.5$~kg), weighing up to 83\% of the robot's weight, into desired positions. Second, we show human-robot collaboration for carrying an object, where the robot and teleoperator take on leader and follower roles. 

Video demo can be found at: https://youtu.be/HalFx296RPo.
\end{abstract}


\IEEEpeerreviewmaketitle

\section{Introduction}
Humanoid robots have the potential to aid workers in demanding manual labor and be general purpose solutions in various industries \cite{humanoid_industry_manufacturing}. To be truly useful and reliable tools these robots must be capable of performing human-like tasks. For example, in manufacturing warehouses, humans coordinate their upper and lower body motion to push heavy boxes into their designated slots. In hospitals, two people work together, providing compliance and safety, to support a patient bed and guide it to a desired location. We qualify the types of skills exhibited in these scenarios as Dynamic Loco-Manipulation (\textbf{DLM}): the simultaneous coordination of locomotion and manipulation to accomplish a forceful physical task that involves the management of dynamic interaction properties (e.g. stiffness, accelerations, contact forces). Teleoperation presents a viable approach to controlling these platforms for DLM by embedding human planning and intelligence into robot control loops. In doing so, we can leverage human adaptability and even attempt to uncover key insights into executing these tasks autonomously.   

\begin{figure}[t]
    \centering
    \includegraphics[width = \columnwidth]{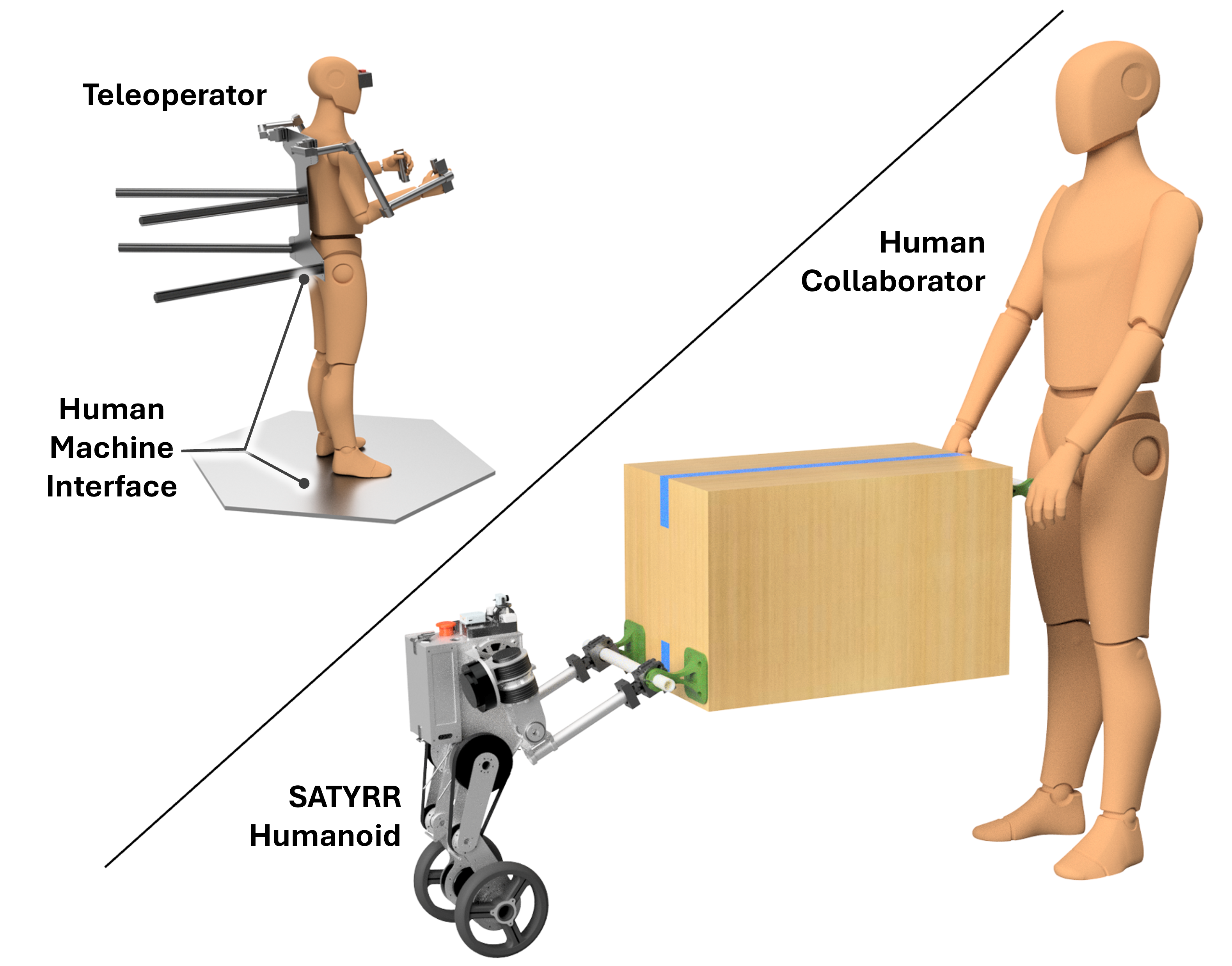}
    \caption{The teleoperator (left) controls whole-body of the robot to accomplish a collaborative carrying task with a human. The human machine interface provides haptic feedback and applies a wrench on the human body, synchronizing the robot and teleoperator body motion.}
    \label{fig:MoneyShot}
    \vspace{-1.75em}
\end{figure}
Our approach enables robot locomotion and manipulation control for DLM via whole-body motion retargeting strategies. In real time, we utilize the humans body motion to control the locomotion of the robot, while the operator can use their arms to directly control robot manipulation \cite{PuruDMM}. Here, we refer to a retargeting strategy or mapping as a function that takes human motion and generates a robot setpoint or reference for a tracking controller. A control mode is a mapping and its corresponding tracking controller. We provide haptic feedback to the pilot's center of mass to enhance their awareness of their environment and create a more immersive teleoperation experience. More specifically, we use a Human-Machine Interface (\textbf{HMI})\cite{wang2021dynamic} to measure human pitch \cite{Puru2022}, capture arm motion, and deliver haptic forces up to 400 N and moments up to 20 N-m. Time-delays are not studied in this work as communication between the robot and HMI run at 200 Hz. 


Previous human-to-robot  retargeting strategies have demonstrated successful tasks such as synchronized bipedal walking \cite{colin2023bipedal, ProfPaper2}, and whole-body control of humanoids \cite{ramos_big_hermes}. Recently, authors in \cite{fu2024humanplus,austin_imitation_learning} used motion retargeting in an imitation learning framework to show successful completion of a number of locomotion and manipulation tasks. \emph{Wu.~et~al} employed predefined switching for disjoint control of locomotion and manipulation \cite{teleop_MOCA}. Authors in ~\cite{icub_teleop_locoman} switch between a higher autonomy mode and a low motion retargeting mode to lift light weight boxes with the iCub robot. However, in these demonstrations the coordination of locomotion and manipulation for forceful tasks or human-robot collaboration (\textbf{HRC}) has yet to be shown. \emph{Portela. et al}, show such forceful interaction by using a quadruped with a mounted arm to pull a heavy box via a learned policy \cite{mit_force_control}. The authors in \cite{hutter_dlm} show a ballbot pushing a heavy door and lifting a heavy item. The authors in \cite{robot_assisted_nav} show loco-manipulation of a mobile manipulator, in an HRC setting, used to guide visually impaired participants to a desired location using an adaptive impedance control strategy.  These works, however, do not leverage teleoperation. Teleoperated human-robot interaction using motion retargeting, and admittance control was demonstrated but solely for the single task of hugging \cite{tHRC_hugging}. Here, we address these issues and show DLM tasks that involve interaction with the environment and HRC. Additionally, we study the efficacy of haptic feedback for enhancing teleoperator awareness about the robot's interaction with the environment and improving DLM task completion. While previous studies have explored force feedback to the pilot's  CoM\cite{colin2023bipedal} or wrist \cite{wrist_haptic}, they have not investigated torsional feedback to enhance understanding of moments at the robot's end-effector and environmental impedance. We propose whole-body moment feedback to enable pilots to adjust their whole-body strategy for DLM.


In our prior work \cite{PuruDMM}, we demonstrated heavy box pushing along a straight path. However, these initial results and mappings were limited. The robot and box were constrained to move in a straight line by railings on the ground. The locomotion retargeting imposed a cognitive burden on operators, requiring them constantly adjust the robot's acceleration for position tracking tasks; furthermore, compliance for human-robot collaboration was not demonstrated. To address these challenges, we enable operators to choose between different position-force locomotion and manipulation modes. Hybrid motion-force control has been studied previosly in applications for manipulator control \cite{hybrid_motion_force_control}. In this work, for locomotion the operator can choose velocity-position control for precise position tracking or acceleration control for dynamic interaction with the environment. For manipulation, the operator can opt for joint-position control for precise arm motion or impedance control for compliant HRC.

The primary contributions of this study are:
\begin{itemize}
    \item A whole-body retargeting framework employing position or force mappings for manipulation and locomotion   
    \item Realization of a whole-body moment feedback for promoting pilot situational awareness of their environment
    \item Experimental validation of the teleoperation framework for DLM tasks such as heavy box slotting and leader-follower style HRC in carrying an object
\end{itemize}

\section{Preliminary}
\label{sec:background}
 Whole-body retargeting strategies for locomotion and manipulation are used to create an intuitive and immersive teleoperation experience. Here, we review two different locomotion strategies and a single manipulation mapping. The pilot uses their body to control the robot's locomotion. The manipulation retargeting aligns the arm motions of human and robot via their respective arm kinematics. Our review highlights these retargeting strategies, discussing their individual strengths and drawbacks.

\subsection{Locomotion Retargeting}
Use of whole-body human motion for hands-free tele-locomotion of humanoids has shown viability \cite{Puru2022,PuruDMM, wang2021dynamic}. Modeling the human as a fixed-base inverted pendulum we can define human pitch, $\theta_H$, as the angle between the human ankle, and their CoM. Modeling the robot as a cart-pole, its linearized dynamics can be written as:
\begin{align}
    \bm{\dot{q}}_{xR} = \bm{Aq}_{xR} + \bm{Bu}_{xR}
\end{align}
where $\bm{q}_{xR} =[x_R\ \theta_R\ \dot{x}_R\ \dot{\theta}_R]^\intercal$ is the position and velocities of the cart's horizontal displacement, $x_R$, along the sagittal plane or robot's $x$ axis, and $\theta_R$ the angle between the cart's base and CoM. 

One retargeting strategy \cite{Puru2022} that enables station keeping and  precise position tracking linearly maps human pitch, $\theta_H$, to the robot cart-pole base velocity, $\theta_H \rightarrow \dot{x}_R^{des}$: 
\begin{align} \label{eq:background_vel_map}
    \dot{x}_R^{des} =  k_{v}\theta_H 
\end{align}
where $k_v \in \R^1$ is a velocity scaling tuned by user preference. Integrating, we can find the desired robot base position, ${x_R^{des} = \int \dot{x}_R^{des} dt}$. Desired pitch, and pitch velocity are set to zero, $\theta_R^{des} = 0$ and $\dot{\theta}_R^{des} = 0$. Consequently, the dynamic responsivity of the robot is limited because the cart-pole model dictates that its acceleration is proportional to its lean, ${\ddot{x}_R \propto \theta_R}$. A similar mapping from human yaw, $\phi_H$, to robot yaw velocity, $\dot{\phi}_R$, can be used to control robot yaw, $\phi_R$.

\begin{figure}[t]
    \centering
    \includegraphics[width = \columnwidth]{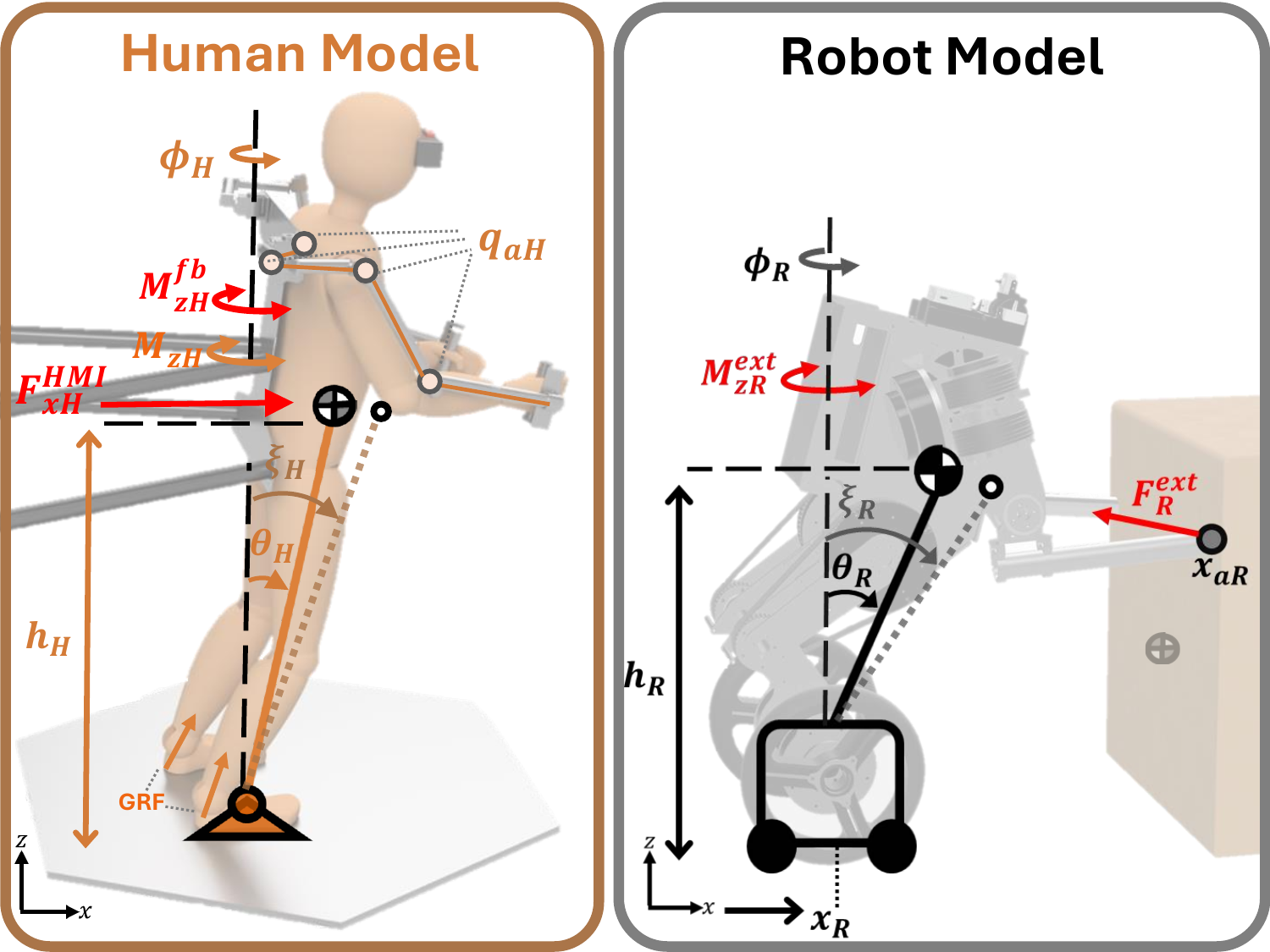}
    \caption{The reduced order models for the human and robot with their corresponding states are shown. The human and robot heights are given by $h_H$ and $h_R$, respectively. Using the forceplate, we are able to measure the ground reaction forces (GRF) and estimate the human's applied moment on the ground, $M_{zH}$.}
    \label{fig:human_robot_rom}
    \vspace{-1.25em}
\end{figure}
  
Conversely, a dynamic locomotion strategy directly maps $(\theta_H,\dot{\theta}_H)$ to $(\theta_R,\dot{\theta}_R)$. The human Divergent Component of Motion (DCM), $\xi_H = \theta_H + \dot{\theta}_H/\omega_H$, is mapped to the robot cart-pole DCM, $\xi_R = \theta_R + \dot{\theta}_R/\omega_R$, accounting for the natural frequencies of the human ($\omega_H = \sqrt{g/h_H}$) and robot (${\omega_R= \sqrt{g/h_R}}$) as shown in our previous work\cite{PuruDMM}:
\begin{align}\label{eq:background_dcm_acc_map}
    \xi_R^{des} = \xi_H 
\end{align}
This strategy is effective for pushing boxes as the pilot can stabilize the robot, directly control it's acceleration, and disable wheel position tracking ($x_R, \dot{x}_R$ are not tracked) to prevent error accumulation and slippage. In turn, the operator must constantly adjust the robot’s acceleration and their pitch to keep the robot in a fixed position, making it difficult to use for precise position control and turning. 

Enforcing this dynamic mapping in \cite{PuruDMM} results in a human-robot normalized haptic feedback force, $F_{xH}^{HMI}$, that attempts to align the the human and robot pitching motion and convey robot contact forces along the $x$ axis, $F_{xR}^{ext}$, to the operator:
\begin{equation} \label{eq:dmm_force_fb}
    F_{xH}^{HMI} = \gamma_H (\xi_R - \xi_H) + \frac{\gamma_H}{\gamma_R}F_{xR}^{ext}
\end{equation} 
where $\gamma_{H}, \gamma_R$ can be viewed as nondimensionalizing scalings for human (H) and robot (R) forces with units Newton. The human-robot states and feedback are shown in Fig.~\ref{fig:human_robot_rom}.

\subsection{Manipulation Retargeting}
To align human and robot arm motion, while accounting for their different sizes, we review a kinematic arm mapping from our previous work \cite{PuruDMM}. 
The human arm joint angles, $\bm{q}_{aH} \in \R^4$, are captured by the HMI exo-suit. Modeling the robot shoulder as a spherical wrist, we use an inverse kinematics (IK) approach to solve for the desired robot joint positions, $\bm{q}_{aR}^{des} = IK_{H\rightarrow R}(\bm{q}_{aH})$. The motor's PD controller tracks the desired joint angles using proportional gain $\bm{K}_p$ and derivative gain $\bm{K}_d$. This mapping enables precise motion tracking of the human arm and renders stiff robot end-effector control.

\section{Method} \label{sec:methods}
Previous work \cite{PuruDMM} demonstrated DLM tasks using acceleration control of the robot's base and precise end-effector control of the arms. In certain scenarios, this approach imposed an undesirable cognitive load on the pilot. For example, during station-keeping, constant adjustments are required to keep the robot stationary. To address this, we enable the pilot to switch between position and force mappings for locomotion and manipulation, allowing for precise control and dynamic interaction with the robot's environment. We introduce a set of input modes for the teleoperator $\mathcal{M} = \{\mathtt{P}, \mathtt{D}\}$ where $\mathtt{P}$ and $\mathtt{D}$ are distinct modes standing for precision or dynamic robot mappings. We do not claim exclusivity in the modes of control for accomplishing DLM; other modes could be used, but we focus our efforts on integrating these two.


Section \ref{method:telelocomotion}, outlines our locomotion retargeting for robot translation and yaw. Section \ref{method:telemanipulation} shows our manipulation retargeting for joint-space or impedance control of the end-effector. Finally, Section \ref{method:momentfeedback} derives the haptic moment feedback at the human torso, that allows the pilot to feel contact moments between the robot and environment.

\subsection{Telelocomotion Retargeting via Position-Force Mappings}
\label{method:telelocomotion}

The telelocomotion retargeting strategy includes mappings: \(\pi_s\) for sagittal plane motion and \(\pi_y\) for yaw rotation around the world frame's \(z\)-axis. To leverage precise position tracking via velocity control (Eq.~\ref{eq:background_vel_map}) and dynamic responsiveness for heavy-box pushing via acceleration control (Eq.~\ref{eq:background_dcm_acc_map}), we define a hybrid sagittal mapping:


\begin{equation}
    \pi_{s}(u_s) :=
    \begin{cases}
        \begin{aligned}
            &\dot{x}_R^{des} = k_v\theta_H &, \quad &u_s = \mathtt{P} \\
             &\xi_R^{des} = \xi_H  &, \quad &u_s = \mathtt{D} 
        \end{aligned}    
    \end{cases}
\end{equation}
where $u_{s} \in \mathcal{M}$ is the human switch input corresponding to precision ($\mathtt{P}$) velocity or dynamic ($\mathtt{D}$) acceleration  control along the sagittal plane. The velocity map scaling, ${k_v = 3.0}$, is tuned by pilot preference for their desired responsiveness and sensitivity. The references ($x_R^{des}, \dot{x}_R^{des}$) and $\xi_R^{des}$ are tracked by LQR controllers $\bm{K}_{vel}$~\cite{Puru2022} and $K_{DCM}$~\cite{PuruDMM}, respectively. These controllers are switched discretely with their corresponding mappings. In a shared-autonomy setting, the switching could also depend on robot states, contact, or user input.



Notably, when switching from acceleration to velocity control, the desired position must be reset so the locomotion controller does not try to correct accumulated position error:
\begin{equation}
            {}^+ x_R^{des} = {}^{-}x_R 
\end{equation}
where $-, +$ describe the pre-transition and post-transition states and setpoint, respectively. 

\begin{figure}[t] 
    \centering{\includegraphics[width=\columnwidth]{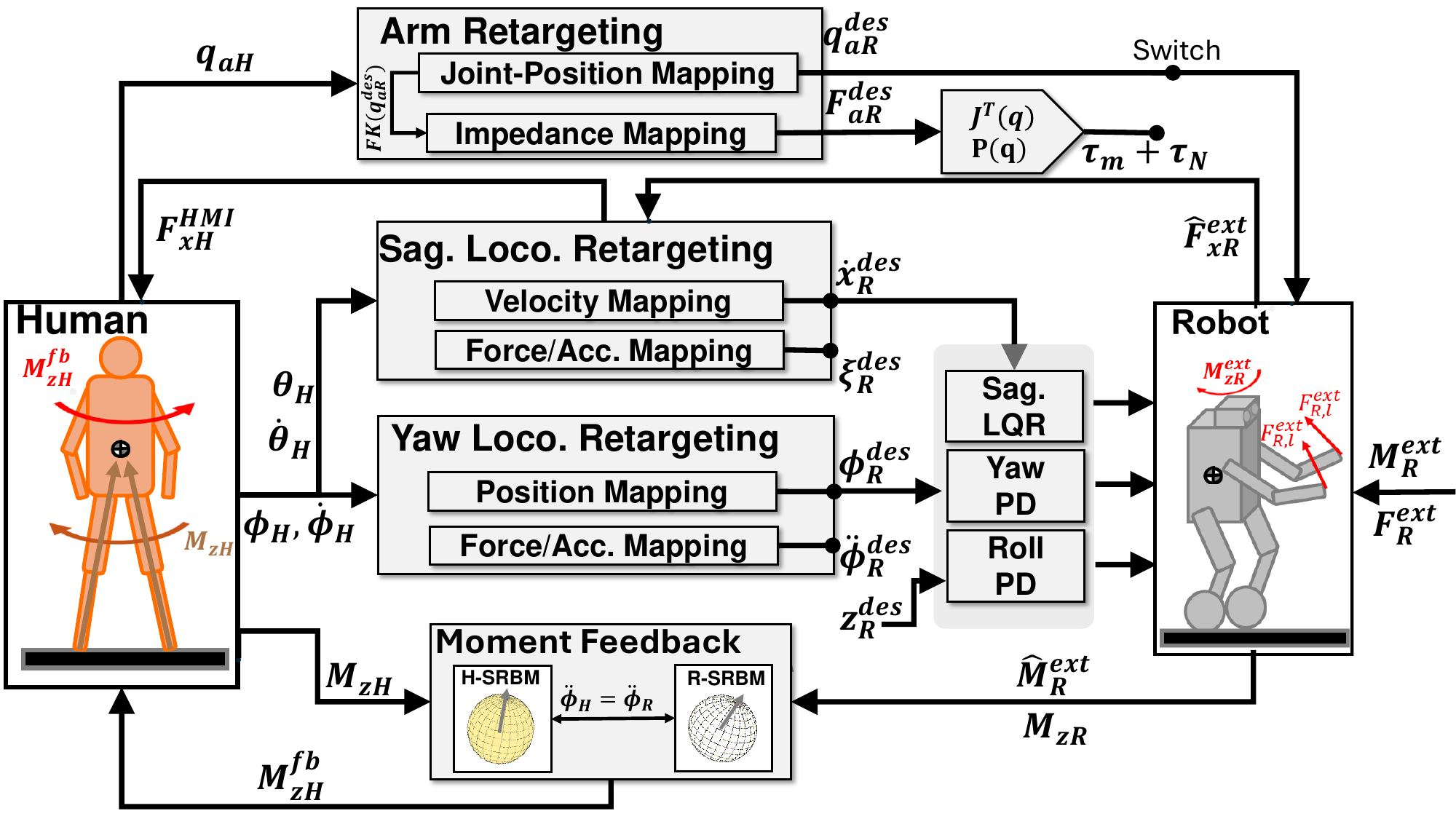}}
    \caption{Full bilateral teleoperation layout and control framework. The pilot chooses 1 of 2 mappings for each of the control modes: sagittal locomotion, yaw locomotion, and arm manipulation. The haptic feedback to the pilot consists of force and moment feedback to synchronize human-robot motion and improve the teleoperator's estimate of the robot's states and of contact with the environment.}
    \label{fig:HMI_wholesys}
    \vspace{-1.25em}
\end{figure}



From our previous experiments \cite{Puru2022,PuruDMM}, we found that yaw velocity control was sufficient for telelocomotion but difficult to manage when pushing heavy boxes due to error in the desired position due to integration of velocity.
Hence, we propose linearly mapping human yaw to robot yaw for intuitive position control:
\begin{equation}
    \begin{aligned}
        &\phi_R^{des} = k_y\phi_H + \phi_{offset} \\
        &\dot{\phi}_R^{des} = k_y\dot{\phi}_H
    \end{aligned}
\end{equation}
where $k_y = 1.5$ increases turning sensitivity and is set by user preference. Here, position control is relative to the yaw offset, $\phi_{offset}^{des}$, that is initialized to zero. While this control mode is intuitive to use, the robot’s yaw is constrained by the pilot’s limited yaw range in the HMI i.e $\phi_H \in [\frac{-\pi}{3}, \frac{\pi}{3}]$. Therefore, we propose an additional mapping that allows the teleoperator to switch to controlling the robot’s angular acceleration and fully rotate the robot as needed:
\begin{align} \label{eq:yaw_acc_map}
    \ddot{\phi}_R^{des} = k_m\phi_H
\end{align}
where \( k_m = 25~ \text{rad/s}^2 \) was tuned by user preference. To achieve the desired angular acceleration and find the corresponding feedforward yaw wheel torque, \( \tau_y \), we model the robot yaw as a differential drive \cite{lynch2017modern} with yaw dynamics:
\begin{equation}
\label{eq:yaw_dynamics_diff_drive}
    I_{zR}\ddot{\phi}_R = dF_R
\end{equation}
where \( d \) is the distance between the robot’s wheels, \( F_R \) is the force on the wheel, and $I_{zR}=0.1$ kg$\cdot$m\raise0.5ex\hbox{2} the robot inertia along $z$ axis. Substituting Eq.~\ref{eq:yaw_acc_map} into Eq.~\ref{eq:yaw_dynamics_diff_drive}, solving for $F_R$, and using the relation between wheel force and torque ($\tau_y = r_w F_R$) we can solve for the desired yaw wheel torque:
\begin{align}
    \tau_y = \frac{r_wI_{zR}}{d}k_m\phi_H
\end{align}
where $r_w$ is the robot wheel radius. The desired robot locomotion is achieved by summing this torque with the translation wheel torque. Therefore, the hybrid yaw retargeting can be defined:
\begin{equation}
    \pi_{y}(u_y) := 
    \begin{cases}
        \begin{aligned}
           &\phi_R^{des} = k_y\phi_H + \phi_{offset} &, \quad &u_y = \mathtt{P}\\
           &\ddot{\phi}_R^{des} = k_m\phi_H &, \quad &u_y = \mathtt{D} 
       \end{aligned}
    \end{cases}
\end{equation}
where $u_y \in \mathcal{M}$ is the human yaw switch input that corresponds to precise ($\mathtt{P}$) yaw position control or dynamic~($\mathtt{D}$) yaw angular acceleration control. The desired yaw position and velocity reference are tracked by a PD controller, whereas the desired angular acceleration is achieved by using the feedforward torque, $\tau_y$. The controllers are switched discretely with their corresponding mappings.


When switching from $\pi_y(\mathtt{D}) \rightarrow \pi_y(\mathtt{P})$, the yaw offset is set to the pre-transition, anticipated robot yaw:
\begin{equation}
    {}^+\phi_{offset} = {}^- \phi_R - k_y {}^- \phi_H
\end{equation}
Through this reset, and by switching between mappings the pilot can control the robot yaw position relative to $\phi_{offset}$, beyond the limited HMI yaw range. The hybrid locomotion mappings are shown in the system block diagram in Fig. \ref{fig:HMI_wholesys}.


 Moreover, since regulating acceleration is proportional to controlling forces and moments, the dynamic interaction modes, $\pi_y(\mathtt{D})$ and $\pi_s(\mathtt{D})$, also provide compliant control of the robot base. Conversely, the precision position and velocity modes, $\pi_y(\mathtt{P})$ and $\pi_s(\mathtt{P})$, offer a stiffer interface, but require less pilot regulation when trying to stay in place or move a long a desired curve. 


\subsection{Telemanipulation Retargeting via Hybrid Mappings}\label{method:telemanipulation}

Similar to the choices for telelocomotion, the teleoperator can select either precision manipulation control or dynamic interaction control of the end-effector for HRC. Joint-space position control offers superior motion tracking of the human arm and a stiff, leader-like interface for the human collaborator in HRC tasks. Contrarily, cartesian space impedance control provides enhanced compliance and safety, for a follower-like interaction with the collaborator. These options provide flexibility in achieving the desired end-effector control. Here, we discuss the Cartesian space impedance controller that also accounts for human-robot arm kinematic differences.

Using the $IK_{H\rightarrow R}$ method outlined in \cite{PuruDMM}, we solve for the desired arm joint angles, $\bm{q}_{aR}^{des}$ that accounts for human-robot kinematic differences. We then use the robot arm's forward kinematics (FK) to find the desired end-effector Cartesian position, $\bm{x}_{aR}^{des} \in \mathbb{R}^3$:
\begin{equation}
    \bm{x}_{aR}^{des} = FK(\bm{q}_{aR}^{des})
\end{equation}

\begin{figure}[t]
    \centering
    \includegraphics[width = \columnwidth]{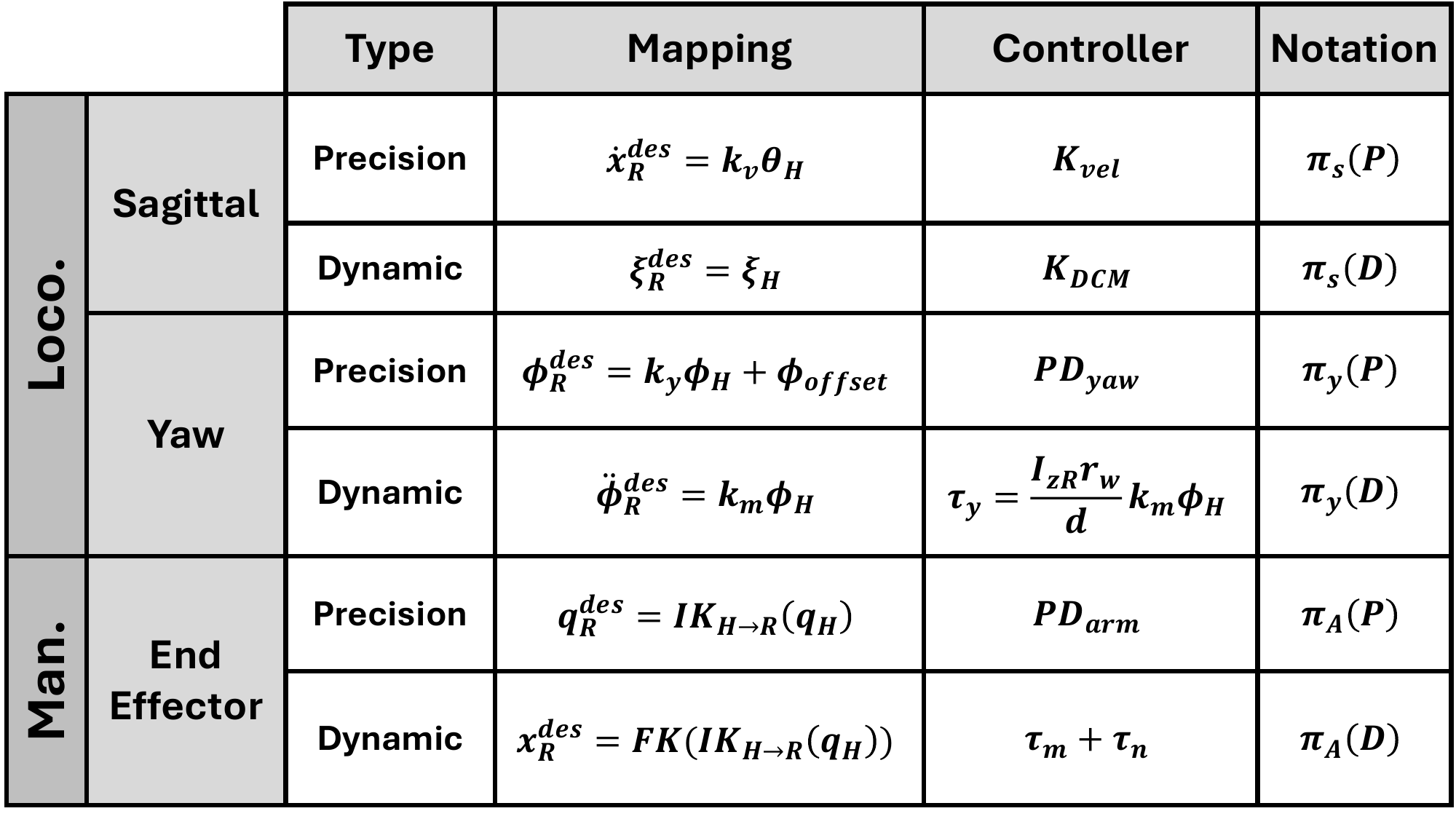}
    \caption{The table summarizes the teleoperator's control modes, highlighting the option to choose between precise position or dynamic force/acceleration mappings for robot locomotion and manipulation.}
    \label{fig:data_table}
    \vspace{-1.5em}
\end{figure}

To introduce compliance and control impedance, we treat the robot end-effector as a 3D spring-damper system:
\begin{equation}
    \bm{F}_s = \bm{K}_{x}(\bm{x}_{aR}^{des} - \bm{x}_{aR}) + \bm{K}_{dx}(\bm{0} - \bm{\dot{x}}_{aR})
\end{equation}
where $\bm{F}_s \in \R^3$ represents the hand virtual-spring force, and $\bm{K}_x \in \R^{3\times 3}$ and $\bm{K}_{dx} \in \R^{3\times 3}$ are positive-definite diagonal matrices that represent the Cartesian stiffness and damping. 

The manipulator dynamics for a single arm are given by:
\begin{subequations} 
    \begin{align}
        \!\!\!\bm{M\ddot{q}}_{aR} \!+\!\bm{C}(\bm{q}_{aR},\bm{\dot{q}}_{aR})\!+\! \bm{G}(\bm{q}_{aR})\! &= \!\bm{\tau}_m\!+\!\bm{\tau}_n\!+\! \bm{\tau}_{ext} \\
         \bm{\tau}_{m} =\, &\bm{J}_{a}^{\intercal} \bm{F}_s \!+\! \bm{G}(\bm{q}_{aR})
    \end{align}
\end{subequations}
Here, $\bm{M}$ is the mass matrix, ${\bm{C}(\bm{q}_{aR},\bm{\dot{q}}_{aR})}$ are the Coriolis and centrifugal forces, and $\bm{G}(\bm{q}_{aR})$ the gravitational forces.  Our applied joint torque, $\bm{\tau}_m \in \R^4$, is the sum of the impedance control torque and gravity compensation. Our arm manipulator Jacobian, $\bm{J}_{a} \in \R^{3 \times 4}$, is rank deficient. To fully control all joints on the arm without interfering with our primary task of emulating a spring-damper, we append a secondary task, $\bm{\tau}_n \in \R^4$, onto the null-space of our Jacobian, ${\bm{P}(\bm{q}_{aR}) \in \R^{4 \times 4}}$: 
\begin{equation}
     \bm{\tau}_N  = \epsilon \bm{P}(\bm{q})[\bm{K}_{p}(\bm{q}^{des}_{aR} - \bm{q}_{aR}) + \bm{K}_{d}(\bm{\dot{q}}^{des}_{aR} - \bm{\dot{q}}_{aR})]
\end{equation}
This task tracks the desired joint angles. To vary the gains of the tracking task we set $\epsilon = 0.15$. 
The teleoperator can then choose between precise or dynamic end-effector control, as defined by the hybrid manipulation mapping:
\begin{equation}
    \!\!\! \pi_{A}(u_A) := 
    \begin{cases}
        \begin{aligned}
            &\bm{q}_{aR}^{des} = IK_{H\rightarrow R}(\bm{q}_{aH}) \!\!\!\!\!&, u_A = \mathtt{P}\\
            &\bm{x}_{aR}^{des} = FK(IK_{H\rightarrow R}(\bm{q}_{aH}))\!\!\!\!\!&, u_A = \mathtt{D} 
       \end{aligned}
    \end{cases}
\end{equation}
where $u_A \in \mathcal{M}$ is the human arm switch input that corresponds to either precise ($\mathtt{P}$) joint position control or dynamic ($\mathtt{D}$) , compliant impedance control. The joint-level PD controller and impedance controller are switched alongside their corresponding mappings. The switching of the manipulation controllers is parameterized by a variable $\alpha(t) \in [0,1]$ that transitions linearly over a span of $70$~ms. The same formulation is mirrored for the other arm.


\subsection{Whole-body Moment Feedback} \label{method:momentfeedback}
To improve situational awareness, we aim to provide the pilot with information about environment impedance and contact moments along the world $z$ axis through haptic moment feedback at the human torso. We model the human-robot rotational dynamics as a single rigid body with roll and yaw given by $\psi$ and $\phi$. Assuming small angular roll velocities (\( \dot{\psi} \approx 0 \)), the yaw dynamics are given by \cite{NASA}:
\begin{equation}
    I_{zj}\ddot{\phi}_{j} = M_{zj}^{net}
\end{equation}
where $j = {H,R}$ represents the human and robot, and $M_{zj}^{net}$ is the net moment on the body. We used the HMI to measure the human's applied moment and angular acceleration, estimating the human's inertia, ${I_{zH}=0.3}$~kg$\cdot$m\raise0.5ex\hbox{2}, offline. The human and robot system of equations are:
\begin{align}
    \vspace{-1.5em}
    \label{eq:sys_of_eq_robot_and_human}
    I_{zR}\ddot{\phi}_{R} &= M_{zR}^{ff} + M_{zR} + M_{zR}^{ext} \\
    I_{zH}\ddot{\phi}_{H} &= M_{zH} + M_{zH}^{fb}
    \vspace{-1.5em}
\end{align}
where $M_{zR}^{ff}$ is a robot feed-forward, $M_{zR}$ is the moment generated on the robot's body by its wheels, $M_{zR}^{ext}$ the moment generated by the contact at the robot's end-effector, $M_{zH}$ the moment generated by the human measured by the forceplate, and $M_{zH}^{fb}$ the moment applied to the human via the HMI. The robot and human angular accelerations and inertia are given as \( \ddot{\phi}_{R}, \ddot{\phi}_{H} \) and \( I_{R}, I_{H} \), respectively.

We aim to equalize the angular accelerations of the human and robot to
(1) explicitly model and map the external moments acting on both systems, (2) ensure synchronized motion upon contact, and (3) account for the size differences between the human and robot:
 \begin{equation}
    \label{eq:desired_ang_acc_robot_vH}
      \ddot{\phi}_R = \ddot{\phi}_{H} 
 \end{equation}
Substituting Eq. \ref{eq:desired_ang_acc_robot_vH} in Eqs.\ref{eq:sys_of_eq_robot_and_human}:
\begin{equation}
\label{eq:dyn_sim_moment_main}
\frac{M_{zH} + M_{zH}^{fb}}{I_{zH}} = \frac{M_{zR}^{ff} + M_{zR} + M_{zR}^{ext}}{I_{zR}}
\end{equation}
Although there are many choices for the feedback and feedforward, we propose a formulation where the external moment and robot induced moment are fed back to the pilot for improved awareness during contact with the environment: 
\begin{equation} \label{eq:moment_fb}
    M_{zH}^{fb} = \frac{I_{zH}}{I_{zR}}(M_{zR} + M_{zR}^{ext})
\end{equation}
To improve responsivity of the robot and satisfy Eq. \ref{eq:dyn_sim_moment_main}, we choose the following feed-forward:
\begin{equation}
    M_{zR}^{ff} = \frac{I_{zR}}{I_{zH}}M_{zH}
\end{equation}
In practice, the feedforward moment was not used (${M_{zR}^{ff} = 0}$). We found that $M_{zH}$ oscillates when the pilot leans forward, resulting in an oscillatory $M_{zR}^{ff}$ and undesired oscillatory robot yaw motion while attempting to move in a straight path. Our initial tests suggest that $M_{zH}$ as a feedforward may not be ideal for wheeled robots, possibly due to natural oscillations of the human ground reaction forces and moments while balancing.

\section{Experimental Setup} \label{sec:exp_setup}
We conducted two DLM experiments. The first experiment extends previous straight box-pushing work to 2D box slotting. The goal was to push a heavy box (40\%–83\% of $m_R$) to two predefined locations. The pilot coordinated the robot's whole-body motion, while adjusting the pushing pose and box contact as needed. Whole-body moment feedback was provided only by user preference. Contact torques greatly exceed inertial torques. Therefore, we estimate the external moment, $\bm{\hat{M}}_{R}^{ext} \in \R^3,$ acting on the robot SRB:
\begin{equation}
    \bm{\hat{M}}_{R}^{ext} = \sum_{i=1}^{n} \bm{r}_{i,j} \times (\lambda \bm{J}_{a,j}^{\dagger}\bm{\tau}_{m,j})
\end{equation}
where $j\in\{r,l\}$ subscripts the left and right hands of the measured motor torques $\bm{\tau_{m,(\cdot)}}$, and the psuedo-inverse of the contact Jacobian, $\bm{J_{a,(\cdot)}}$. The contact variable, $\lambda$, is updated based on contact sensors located at the end of each hand. These sensors consist of an array of tactile buttons that enable sensing in 3D across all three directions: $x, y,$ and $z$. To smooth contact switching, the state machine must remain in one of four states (no contact, right-hand contact, left-hand contact, or both hands contact) for at least 55 ms. To enhance the pilot's perception of environmental contact and account for estimation errors in human inertia, we scale the contact moment feedback based on pilot preference for an optimal teleoperation experience:
\begin{equation}
    \bm{\hat{M}}_{R}^{ext} \leftarrow K_{fb}\bm{\hat{M}}_{R}^{ext}
\end{equation}
where $K_{fb} = 0.5$. Along the sagittal plane, the external contact feedback force in Eq. \ref{eq:dmm_force_fb} was set to zero, ${\frac{\gamma_H}{\gamma_R}F_{xR}^{ext} = 0}$, to isolate and explore the effectiveness of contact moment feedback. Successful completion of these experiments required pushing the box into the predefined slot and against the backrest or wall. All other runs were deemed unsuccessful. 



In the second set of experiments, we demonstrate a HRC task of carrying a lightweight object ($<1$ kg). First, the teleoperator takes a follower role, allowing the human collaborator to guide the object. Compliance at the robot base and end-effector was desired for safety and control. Next, the teleoperator acts as the leader, guiding the object and collaborator. The robot and human convey intent through forces and moments on the object. This experiment highlights the importance of whole-body position-force control modes for DLM in an HRC setting.

\section{Results \& Discussion} \label{sec:results}

Section \ref{Exp:box-slotting-result} presents experiments demonstrating box slotting as a form of DLM. Section \ref{Exp:HtRc} showcases pHRI in a collaborative object-carrying task, emphasizing compliance as crucial for DLM in HRC. Section \ref{Exp:Moment-feedback-result} evaluates the effectiveness of whole-body moment feedback in augmenting teleoperation for DLM. The experiments comply with UIUC Internal Review Board (IRB) requirements and are validated on video: https://youtu.be/HalFx296RPo

\subsection{Box Slotting}
\label{Exp:box-slotting-result}
Triggers located on the right and left hands of the HMI upper body joysticks enabled switching between the sagittal plane and yaw locomotion mappings, $\pi_{s}$ and $\pi_{y}$. The binary trigger at each hand takes values $0$ or $1$ corresponding to input modes $\mathtt{P}$ or $\mathtt{D}$, respectively. In these experiments the boxes weighed between ${5-10.5}$ kg (${40\%-83\%}$ of $m_R$). 

\begin{figure*}[t] 
    \centerline{\includegraphics[width=17.5cm]{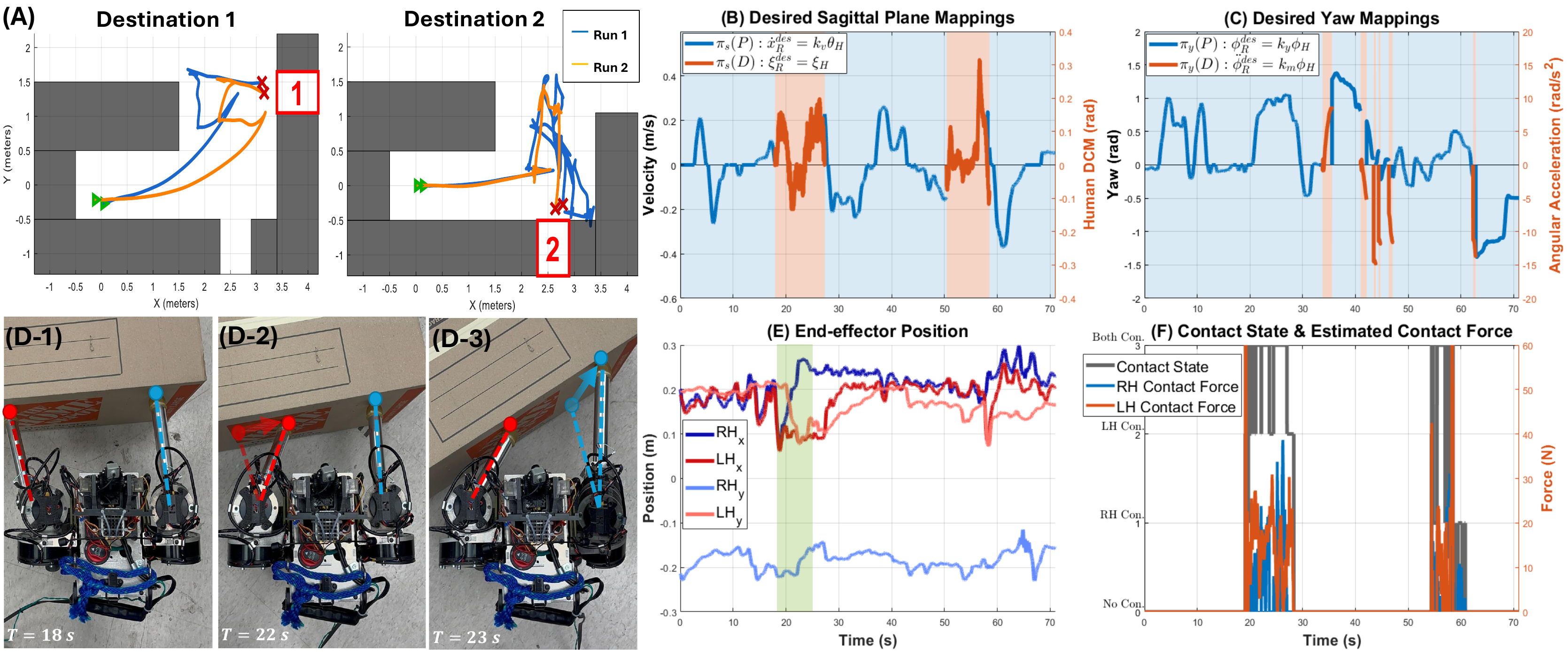}}
    \caption{(A) displays the robot odometry for two box-pushing trials with different end locations: Destination 1 and 2. (B) shows the sagittal locomotion switching behavior where the pilot uses precise velocity mode (in blue) to align the robot pre-contact, and the dynamic acceleration mode (in orange) while pushing. (C) shows the yaw locomotion modes used and highlights the pilots preference for precise position control (in blue) during the majority of the task. The graphic in (D) and the green area showing the arm $(x,y)$ position in (E), highlight the pilot strategy for rotating the heavy-box by moving the left and right hands. (F) shows the contact forces at each hand and the rapid changes in contact as sensed by the contact sensor at each hand.}
    \label{fig:BoxPushData}
    \vspace{-1.25em}
\end{figure*}

At the start of the task, from $t=0 - 18$~s, the pilot utilizes precision sagittal locomotion control, $\pi_s(\mathtt{P})$ , to align the robot with the box without having to adjust acceleration and correct for position drift, as shown in Fig.~\ref{fig:BoxPushData}~(B). This pre-contact alignment helps push the box on a curved path by positioning the robot off-center from the box’s midpoint. From ${t=18-27}$~s, the pilot switches to $\pi_{s}(\mathtt{D})$ to dynamically control the robot's DCM, accelerate its base, and disable wheel position tracking. In conjunction, the pilot uses $\pi_{y}(\mathtt{P})$ from ${t=0-33}$~s, since only forward movement at a slight angle - without full rotation - is needed. To break stiction and initiate the box's motion, the pilot uses the robot’s arms to apply an upward lift on the box, reducing normal and frictional forces. Figure~\ref{fig:BoxPushData}(A) shows the robot's odometry and paths taken during two box-pushing trials for each end destination.

Here, we examine a strategy the pilot employed for adjusting the box's yaw. This method involved varying the distance of the contact point from the box's center, as illustrated in Fig.~\ref{fig:BoxPushData} (D). To control applied moments (${M = r \times F}$), the pilot can adjust either the moment arm length ($r$) or the applied forces ($F$). At $t = 19.61$~s, as shown in the highlighted region in Fig. \ref{fig:BoxPushData} (E), the pilot decreases the left hand's y-position (LHy) while pushing outward with the right hand (increasing RHx), effectively rotating the box. By altering the y contact position, the pilot changes the lever arm relative to the box's center of mass, modulating the applied moment. This may also explain the pilot's preference for using joint-space control, $\pi_A (\mathtt{P})$, for precise motion tracking during all box-slotting trials. These initial experiments suggest that the teleoperator adjusted $r$ due to the precise and reliable motion control of the end-effector. This approach may be advantageous compared to precise regulation of contact forces because the estimates of $F$ and the resulting haptic feedback are volatile and noisy due to rapid changes in contact when sliding a box. Figure \ref{fig:BoxPushData} (F) shows the right hand frequently losing contact while pushing to destination 1, and forces on the right hand changing between $0-30$~N. This observation aligns with cognitive science findings that humans integrate visual and haptic information based on the most reliable sensory channel \cite{human_haptic_vision_fusion}. Our experiments suggest that the pilot could employ various strategies to achieve the desired box motion, such as lifting more on one side of the box. These emergent behaviors highlight the teleoperator's adaptability in coordinating the robot's upper and lower body motion to guide the box to its desired destination within the whole-body teleoperation framework.


\begin{figure}[t]
    \centering
    \includegraphics[width = \columnwidth]{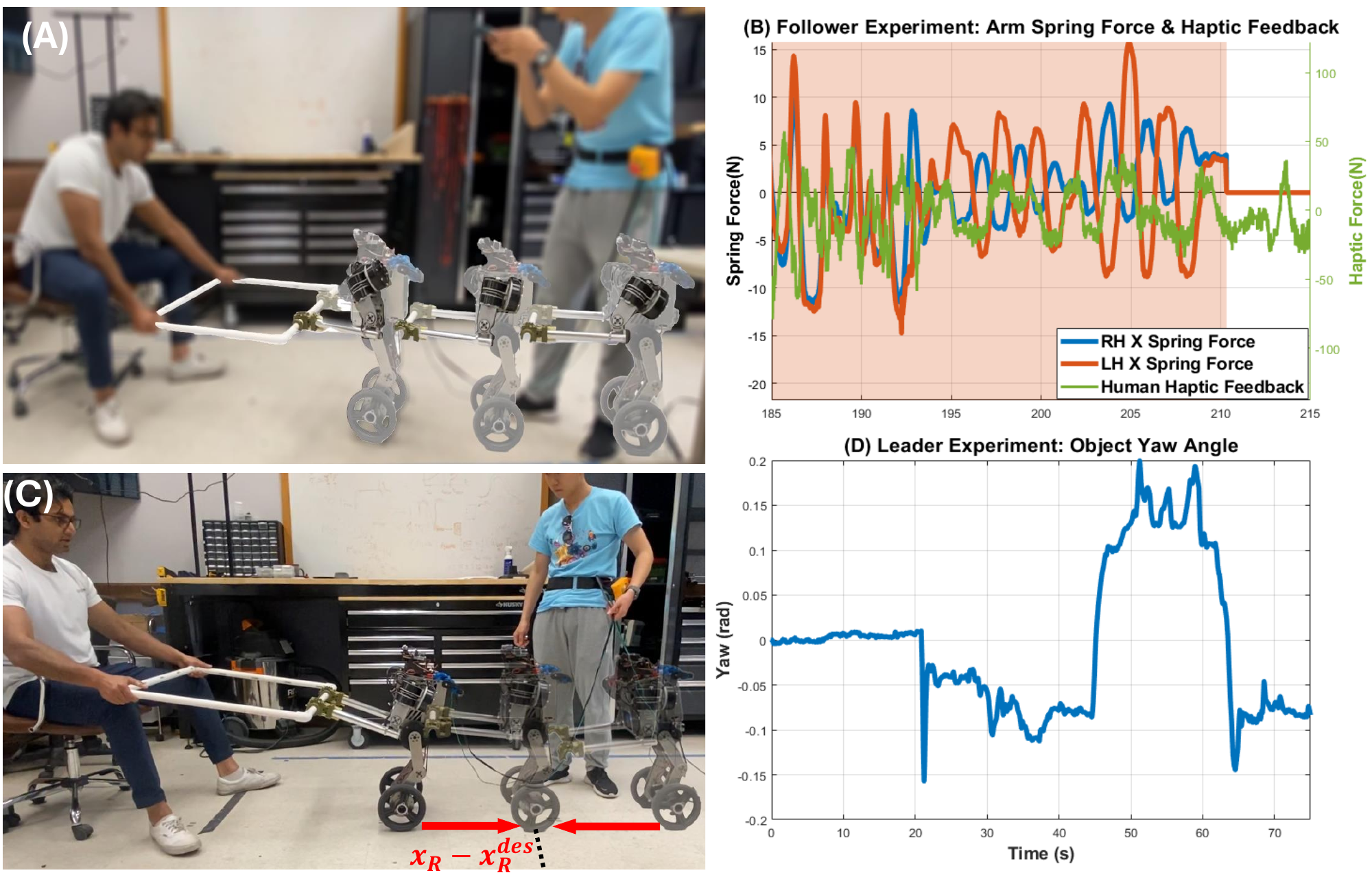}
    \caption{(A) shows the robot standing upright with the arms and base compliant, responding to the collaborator's commands. (B) shows the spring forces at each hand and the force feedback to the pilot. (C) highlights the robot's undesirable pushing/pulling behavior when using precise position control instead of dynamic and compliant modes for HRC. (D) shows the bar rotation enabled by impedance control of the end-effector}
    \label{fig:hri}
    \vspace{-1.25em}
\end{figure}

Next, the pilot breaks contact and switches both locomotion mappings mappings, ($\pi_{s}$) and ($\pi_{y}$), to rotate and readjust the robot's pose. Quick switching from ${\pi_{y}(\mathtt{D}) \rightarrow \pi_{y}(\mathtt{P})}$ stops any undesired rotation. Finally, the pilot applies a push using the previously outlined strategy to slot the box into place. Once the box stops moving, the pilot ceases pushing, successfully completing the slotting task. In summary, Fig.~\ref{fig:BoxPushData}~(B),~(C) show that for heavy box slotting, the pilot prefers sagittal plane velocity control when not in contact, sagittal plane acceleration control during contact, and yaw position control most of the time, except when rotating the robot's body beyond the teleoperator's limited workspace.



\subsection{Human-Robot Collaboration via Teleoperation}
\label{Exp:HtRc}

In these experiments, we focus on stiffness and compliance as they are key in establishing leader or follower roles in carrying objects. By switching between impedance control, $\pi_A(\mathtt{D})$, and joint-space control, $\pi_A(\mathtt{P})$, we change the stiffness at the end-effector for these HRC tasks. The left trigger switches manipulation mappings, while the right trigger switches locomotion mappings. The triggers toggle between input modes $\mathtt{P}=0$ and $\mathtt{D}=1$.

In the first set of trials the teleoperator takes a follower role by trying to comply with the collaborators desired motion. To begin, the pilot is asked to approach one side of the object and latch on. To do so, the pilot uses $\pi_A(\mathtt{P})$ for precise motion control of the arms. The base is controlled using velocity control, $\pi_s(\mathtt{P})$, so that the pilot can focus on the task of approaching, slowing down, and grabbing onto the pipe without having to regulate the acceleration. Once latched on, the pilot switches into acceleration control, $\pi_s(\mathtt{D})$, of the base and impedance control of the arms, $\pi_A(\mathtt{D})$. This allows the human collaborator to easily translate and rotate the robot. 
Since the teleoperator directly controls the robot's acceleration rather than position, the locomotion mappings provide compliance at the base. In Fig. \ref{fig:hri} (A) and in Fig.~\ref{fig:hri}~(B) from ${t = 185-195}$~s, the collaborator moves the robot forward and back while the robot remains upright without tracking position. At this time the robot arms apply synchronized spring forces in response to the collaborator's motion. Solely using velocity control at the base requires the pilot to regulate the desired position ($x_R^{des} = \int \dot{x}_R^{des}dt$), which can lead to the robot base pulling or pushing on the collaborator undesirably if there is an error in the teleoperator's estimate of where the robot should be as shown in Fig. \ref{fig:hri} (C). In the extreme case when there is drift of the desired velocity, the robot can lean too far and loose balance. The low-stiffness arm impedance control enables the collaborator to rotate the object in place without the teleoperator's adjustment of the robot arms. From $t = 195-210$~s, in Fig. \ref{fig:hri} (B), the collaborator rotates the object clockwise and counterclockwise while the robot body remains stationary. Here, each hand applies an opposite spring force to provide compliance in response to the collaborators motion of the object. Contrarily, while using $\pi_A(\mathtt{P})$ the collaborator experiences a stiff interaction and relies solely on the teleoperator's judgment to adjust the robot arms. Through the acceleration and impedance retargeting strategies, the robot demonstrates whole-body compliance for HRC as shown in Fig. \ref{fig:hri}(A).

\begin{figure}[t]
    \centering
    \includegraphics[width = \columnwidth]{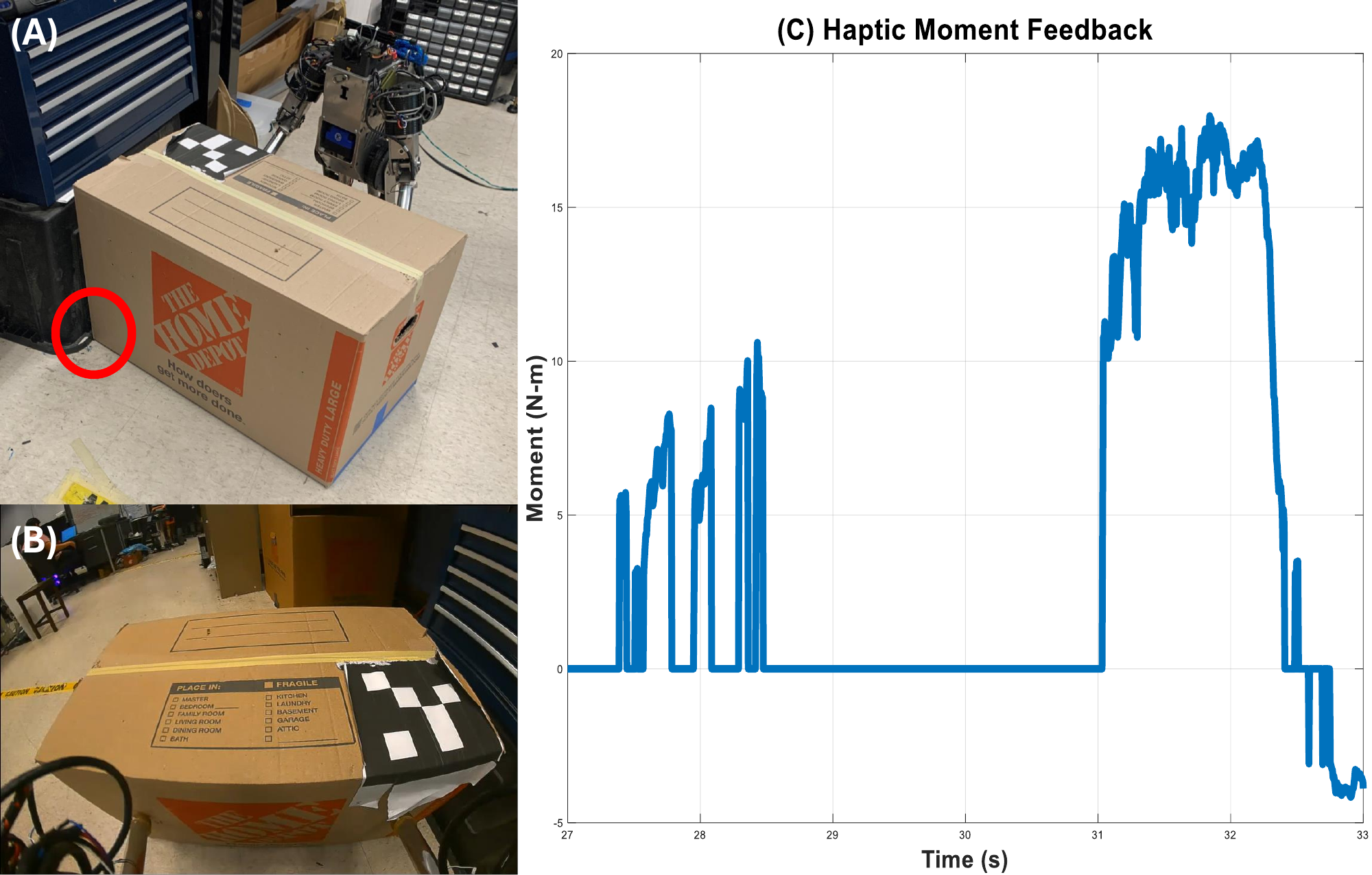}
    \caption{A use case for moment feedback is shown when the pilot is unable to discern why the box is stuck. The contact is occluded from the pilot point-of-view. Haptic feedback here enables the pilot to understand how hard they are pushing. In an attempt to dislodge the box, the pilot pushes with their right hand inducing a contact moment between $t=31-32$~s.}
    \label{fig:BoxStuck}
    \vspace{-1.75em}
\end{figure}

In the second set of trials, the teleoperator was given a leader role and asked to guide the collaborator. Here the pilot used mappings $\pi_{s}(\mathtt{P})$ and $\pi_{A}(\mathtt{P})$ for a majority of the time. The teleoperator commands a forward velocity and pushes with their arms to guide the collaborator along an arc. At times, the teleoperator would switch into $\pi_{s}(\mathtt{D})$ to reset the desired base position, zero the integrated desired position error, and prevent wheel slippage if they noticed the robot was leaning too far or the human collaborator was resisting motion. To rotate the object and collaborator, as shown at $t=44$~s in Fig. \ref{fig:hri}(D), the teleoperator pulls one of the robot’s hands back and the other forward. As the joint-level control renders stiff and precise end effector control, the robot can apply larger forces and moments through the bar at the teleoperator's command. Following this strategy, the teleoperator and robot can lead the collaborator along an arc. The following experiments highlight the effectiveness of hybrid teleoperation mappings for DLM.

\subsection{Evaluating use case of moment feedback}
\label{Exp:Moment-feedback-result}

Whole-body force and moment feedback play a nuanced role in these experiments. In practice, we provided the pilot with contact feedback only, $M_{zH}^{ext}$, in Eq. \ref{eq:moment_fb}. This decision was made because human perception struggles to detect smaller haptic signals at their torso ($M_R$ is small). Additionally, when pushing heavy boxes, the forces and moments from contact greatly outweigh the inertial moments ${(M_{zH}^{ext} >> M_{zR})}$. Thus, we approximate ${M_{zH}^{fb} \approx \frac{I_{zH}}{I_{zR}}M_{zR}^{ext}}$.

For most box-slotting and all table-carrying HRC trials, the pilot preferred $M_{fb} = 0$, relying solely on the dynamic force feedback $F_{xH}^{HMI} = \frac{\gamma_H}{\gamma_R}(\xi_H - \xi_R)$ \cite{PuruDMM}. This feedback enhanced the sensation of being pushed and pulled by the collaborator and helped distinguish between drifting and intentional movements. In our HRC experiments, the moment feedback was not useful for task completion. During the follower experiment, $M_{zH}^{fb}$ did give the pilot a sensation of the impedance forces and moments at the end-effector as the bar was rotated. However, the pilot's coordination with the human collaborator achieved similar task completion results using only visual feedback and FPV view. In the leader experiment, the human collaborator quickly complied with the robot's desired motion through visual cues, resulting in smaller contact moments. Again, the teleoperator achieved similar task completion with visual cues alone. While pushing the box, the estimated moment changed rapidly, as shown in Fig. \ref{fig:BoxPushData} (D). This jerky estimation likely resulted from the box intermittently sliding and stopping and the teleoperator moving the robot’s contact hands, causing repeated making and breaking of contact. This volatility made it difficult for the pilot to make intelligent planning decisions from the haptic channel.


The pilot requested feedback situationally, during test runs when the box got stuck while slotting. In Fig.~\ref{fig:BoxStuck}~(A)-(B), the back right corner of the box hits an external object, preventing rotation, and the pilot cannot see the contact due to occlusion. The pilot could not move the box using visual feedback alone and was unsure if or how much force was being applied. Moment feedback provided the pilot with a sense of the force applied by each hand, enabling them to decide if the robot was applying significant force. Confirming increased moment feedback at $t = 31$s in Fig. \ref{fig:BoxStuck}~(C), the pilot tried a different strategy, focusing on lifting the box. Since the box was not sliding, the contact with the box was fixed, and the estimated external moment feedback was smoother. Using both visual and haptic feedback, the pilot was able to dislodge the box. Moment feedback was then disabled, and the pilot continued the slotting task.

\vspace{-0.5em}
\section{Conclusion} \label{sec:conclusion}

In this paper, we studied the efficacy of hybrid switched mappings and moment feedback for dynamic loco-manipulation with a wheeled humanoid. In the box slotting experiment, the moment feedback allows the pilot to sense the magnitude of force applied to the object but is only enabled upon the pilot’s request while the box was stuck. Position control facilitates more precise operation of the arm and adjustment of the robot's yaw angle. Dynamic force control for sagittal plane motion and yawing allows the pilot to push heavy boxes, fully rotate the robot, and add compliance. The choice of controller is left to the pilot's discretion under different operating conditions. In the human-robot collaboration experiments, position control enables the pilot to operate the robot with precise movements and take a leader role, while impedance control and acceleration control of the base allows provides whole-body compliance for follower role. Future works look to examine the conditions for switching between different control modes, and building further use cases for the moment feedback.


\appendices
\bibliographystyle{IEEEtran}
\bibliography{Main.bib}
\end{document}